\documentclass[sigconf]{acmart}
\AtBeginDocument{%
  \providecommand\BibTeX{{%
    \normalfont B\kern-0.5em{\scshape i\kern-0.25em b}\kern-0.8em\TeX}}}

\setcopyright{acmcopyright}

\copyrightyear{2023}
\acmYear{2023}
\setcopyright{acmlicensed}\acmConference[SIGMOD-Companion '23]{Companion of the 2023 International Conference on Management of Data}{June 18--23, 2023}{Seattle, WA, USA}
\acmBooktitle{Companion of the 2023 International Conference on Management of Data (SIGMOD-Companion '23), June 18--23, 2023, Seattle, WA, USA}
\acmPrice{15.00}
\acmDOI{10.1145/3555041.3589672}
\acmISBN{978-1-4503-9507-6/23/06}

\usepackage{xspace}
\usepackage{pifont}

\definecolor{afblue}{rgb}{0.36, 0.54, 0.66}

\begin{document}

\title{Growing and Serving Large Open-domain Knowledge Graphs}
 


\author{Ihab F. Ilyas}
\email{iilyas@apple.com}
\affiliation{
	\institution{Apple}
	\city{}
	\country{}
	}

 \author{JP Lacerda}
\email{jpl@apple.com} 
\affiliation{
	\institution{Apple}
	\city{}
	\country{}
	}

 \author{Yunyao Li}
\email{yunyaoli@apple.com} 
\affiliation{
	\institution{Apple}
	\city{}
	\country{}
	}

 \author{Umar Farooq Minhas}
\email{ufminhas@apple.com} 
\affiliation{
	\institution{Apple}
	\city{}
	\country{}
	}

  \author{Ali Mousavi}
\email{amousavi@apple.com} 
\affiliation{
	\institution{Apple}
	\city{}
	\country{}
	}

  \author{Jeffrey Pound}
\email{pound@apple.com} 
\affiliation{
	\institution{Apple}
	\city{}
	\country{}
	}

  \author{Theodoros Rekatsinas}
\email{trekatsinas@apple.com} 
\affiliation{
	\institution{Apple}
	\city{}
	\country{}
	}

   \author{Chiraag Sumanth}
\email{csumanth@apple.com} 
\affiliation{
	\institution{Apple}
	\city{}
	\country{}
	}
\renewcommand{\shortauthors}{Ihab F. Ilyas et al.}

\begin{abstract}
Applications of large open-domain knowledge graphs (KGs) to real-world problems pose many unique challenges. In this paper, we present extensions to Saga~\cite{saga} our platform for continuous construction and serving of knowledge at scale. In particular, we describe a pipeline for training knowledge graph embeddings that powers key capabilities such as fact ranking, fact verification, a related entities service, and support for entity linking. We then describe how our platform, including graph embeddings, can be leveraged to create a Semantic Annotation service that links unstructured Web documents to entities in our KG. Semantic annotation of the Web effectively expands our knowledge graph with edges to open-domain Web content which can be used in various search and ranking problems. Finally, we leverage annotated Web documents to drive Open-domain Knowledge Extraction. This targeted extraction framework identifies important coverage issues in the KG, then finds relevant data sources for target entities on the Web and extracts missing information to enrich the KG.

Finally, we describe adaptations to our knowledge platform needed to construct and serve private personal knowledge on-device. This includes private incremental KG construction, cross-device knowledge sync, and global knowledge enrichment.
\end{abstract}

\maketitle

\section{Introduction} 

\begin{figure*}
    \centering
    \includegraphics[width=.8\textwidth]{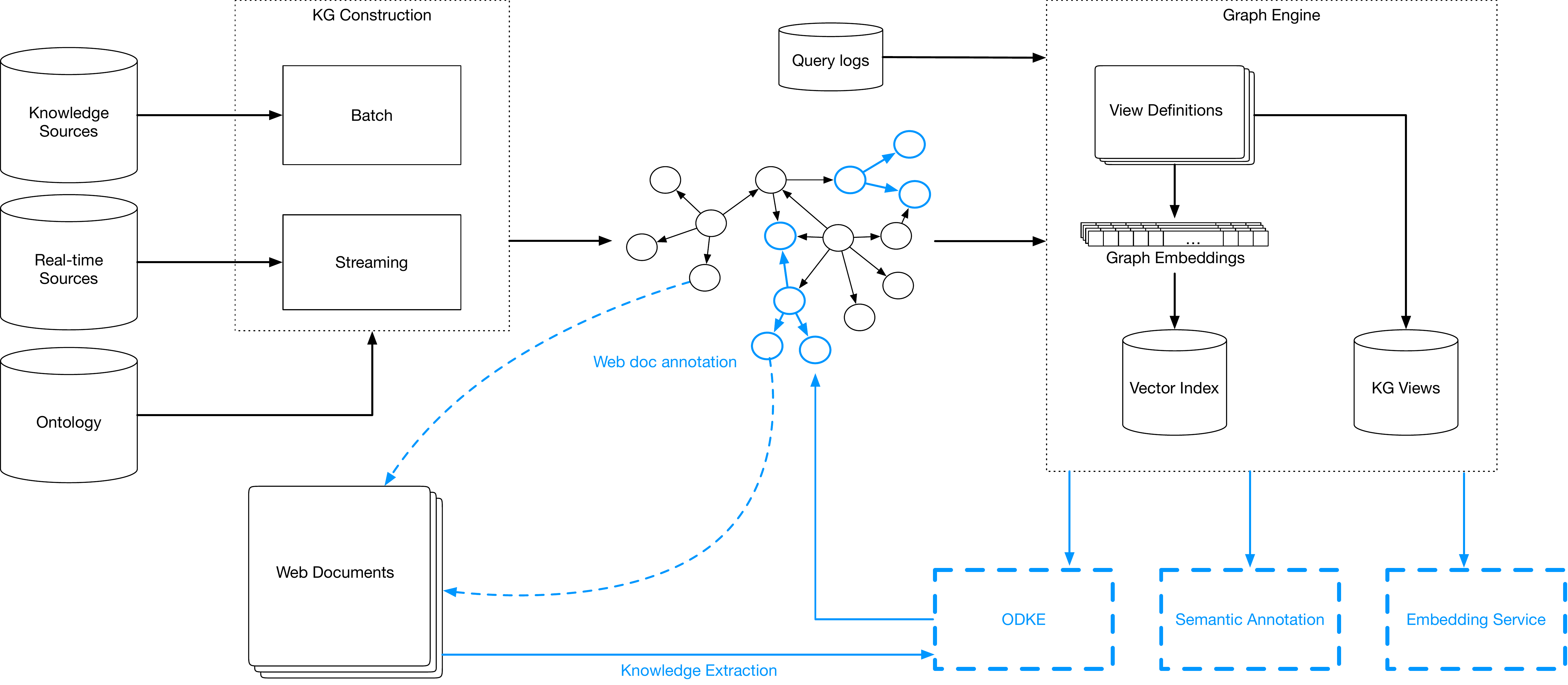}
    \caption{The Saga Knowledge Platform extended for Graph Embeddings, Semantic Annotation for Linking Web Documents to KG entities, and   Open Domain Knowledge Extraction (ODKE).}
    \label{fig:arch}
\end{figure*}

Constructing and serving large knowledge graphs (KG) has become a common component of many important real-world application scenarios such as search, question answering, entity linking, and information extraction. In these use cases, knowledge from a KG is used to construct enriched representations of queries, documents, and entities in order to support semantic reasoning grounded in factual knowledge. As an example, a query \textit{``benicio del toro movies''} can be semantically annotated with:
\begin{center}
\textit{``benicio del toro''} $\rightarrow$ \texttt{entityid\_123}, \\
\textit{``movies''} $\rightarrow$ \texttt{ontology\_type\_movie}
\end{center}
to support retrieval of matching entities from a KG (ie., movies directed by Benicio Del Toro); ranking of facts from a KG (ie., which movies directed by Benicio Del Toro are most important); retrieval and ranking of Web documents that mention Benicio Del Toro and his movies; and related entities (e.g., other similar movie directors) which can be used to suggest related queries to a user.

We previously introduced Saga~\cite{saga}, a state-of-the-art platform for constructing and serving knowledge at scale. In this work, we describe applications of and extensions to Saga which leverage ML-based KG techniques for an important set of problems related to growing and serving knowledge graphs. In particular:
\begin{itemize}
    \item We present a graph embedding pipeline built on Saga's Graph Query Engine that \emph{trains and indexes embeddings of KG entities}. These graph embeddings have many direct use cases, such as fact ranking, fact verification, entity linking, and computing related entities. 
    
    \item We describe a platform for scalable semantic annotation of text. In particular, we address challenges in scaling semantic annotation to tackle the problem of \emph{linking the Web}; to annotate occurrences of entities occurring in a large corpus of Web documents.

    \item Many KG use cases depend on good coverage of facts and entities. To improve KG coverage, we describe our approach to Open Domain Knowledge Extraction (ODKE), a targeted information extraction technique designed to extend the KG with high-valued facts and entities.

    \item Finally, we discuss challenges in applying our platform on-device. In this scenario, data is private and all computation must happen locally on the device. Constructing and serving a KG in a resource constrained environment poses a number of unique challenges; including privacy, resource constraints, cross-device sync,
    and global knowledge enrichment.
\end{itemize}

\paragraph{Overview}
Figure~\ref{fig:arch} gives an overview of our extensions to Saga. An embedding service provides access to learned vectorized representations of entities, and allows similarity calculations as well as efficient $k$-nearest-neighbour retrieval. Next, our Semantic Annotation service leverages graph embeddings to link mentions of entities in text to the knowledge graph. We apply this service at scale to ``link the Web'', extending our KG with edges linking KG entities to unstructured Web documents. Finally, we leverage annotated Web documents as well as traditional retrieval methods to find sources of missing KG information. We then extract missing KG facts, improving coverage of our KG's data.


    


\section{Knowledge Graph Embeddings} 
\label{sec:embeddings}

  Knowledge graphs (KGs) can be instrumental to many downstream machine learning (ML) applications in virtual assistants. Examples of these applications include:
\begin{itemize}
    \item \textit{Fact Verification}. Industrial-scale KGs are continuously updated based on new data from different sources and across diverse domains. Hence, it is necessary to reason about the correctness and completeness of these facts at scale.
    \item \textit{Fact Ranking}. Entities in KGs might have several facts associated to a specific relation. As an example, a person might have several occupations noted in a KG. Therefore, for a query such as \textit{"What is the occupation of X?"} a virtual assistant needs to infer an importance-based ranking over facts in the graph to generate a high-quality answer.
    \item \textit{Related Entities}. When virtual assistants are queried about a specific entity, they can help users to discover facts and learn about its related entities. This pro-active way of information representation provides a richer user experience. 
    \item \textit{Entity Linking}. In order to give a proper answer to a query, or sometimes a richer answer, virtual assistants need to identify the set of KG entities that are present in the user query and the corresponding answer. Therefore, they should be able to detect mentions and properly link them to entities in KGs.  
\end{itemize}

\begin{figure}[t]
    \centering
    \includegraphics[width=0.5\textwidth]{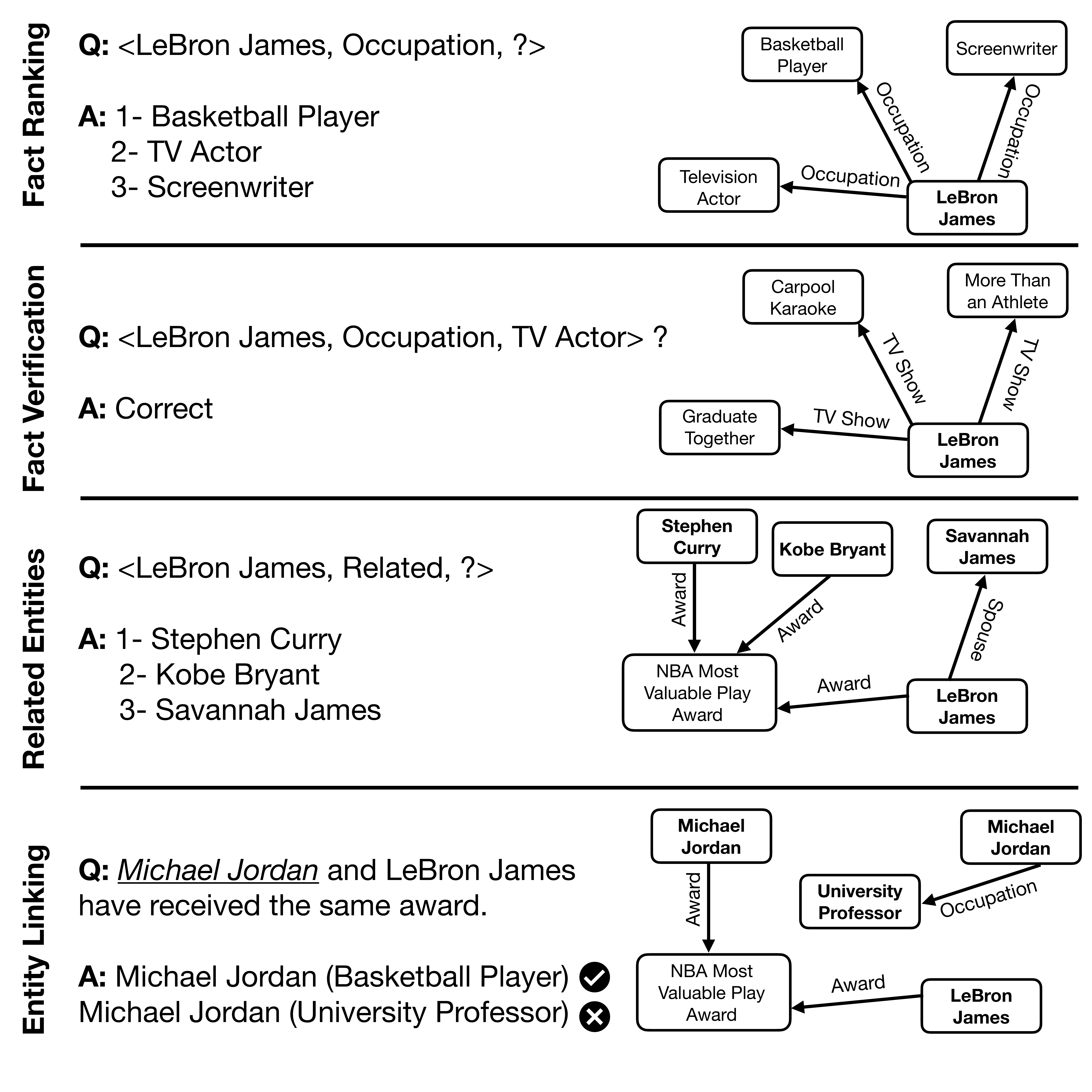}
    \caption{Examples of machine learning applications for virtual assistants in which KGs and their embeddings play a central role.}
    \label{fig:embed_tasks}
\end{figure}

Figure \ref{fig:embed_tasks} depicts example queries and answers for each of these applications. In all of these applications, we have an ML system which uses a KG as input during reasoning. Therefore, a major success factor in designing such systems is to represent a KG in a space that is easy to consume by ML models. Given that most modern ML models operate over vector-based representations, we represent the entities and predicates of a KG using a continuous vector space such that their inherent structure is preserved. KG representation learning comes with its own challenges such as \textit{vectors being trained on non-relevant or noisy data that may exist in the KG, scalability of training and inference, and the design of the embedding model}. 

Open-domain knowledge graphs include data from different sources. Beyond exhibiting a rich semantic structure (consider the union of multiple schemata), it is also possible that facts in the KG are noisy. In this setting, many facts can be irrelevant to a specific downstream embedding task. For example, facts that capture numeric values of entities might not be useful for generating embeddings useful for a related entities service. Example of such facts can correspond to the height of a person, or the National Library ID of a book, or the number of social media followers of celebrities, which while useful for question answering might not be important for learning an embedding for an entity. As a result, such facts may need to be ignored during the training process. This process is also common in fine-tuning foundation models if one wants to achieve high accuracy for specific downstream tasks~\cite{xie2023}.

After filtering non-relevant facts, frequency of certain predicates in the graph could drop to below a threshold and we might not have enough information about those predicates to learn high-quality representations for them. Therefore, triples having those predicates could create noise during the learning process and filtering them out can produce a cleaner training set. 

Industrial-scale KGs can scale to billions of facts and entities. Therefore, scalability of training methods is one of the major issues to consider for learning high-quality representations. At a high-level, we consider two major types of KG embedding models~\cite{HamiltonYL17}: 1) shallow embedding models and 2) reasoning-based embedding models. For shallow embedding models, random edge-based partitioning of the graph is a major technique to combat the scalability challenge and hence, they can easily benefit from multi-node distributed training. Shallow embedding models often learn embedding matrices of entities and predicates by optimizing a contrastive objective on both existing and non-existing edges in the graph. On the other hand, reasoning-based embedding models are used for more complex tasks that involve multi-hop reasoning combined with logical operators. In this case, edge-based graph partitioning can yield poor locality for graph neighborhoods and thus different approaches are needed to obtain high-quality models. There is a diverse array of solutions to address this challenge: 1) many systems opt for infrastructure with sufficient main or GPU memory to hold the entire graph, 2) others use IO-optimized disk-based graph operations to enable global sampling of graph neighborhoods, and 3) several propose pre-computing graph traversals to constract samples.

\begin{figure}[t]
    \centering
    \includegraphics[width=0.5\textwidth]{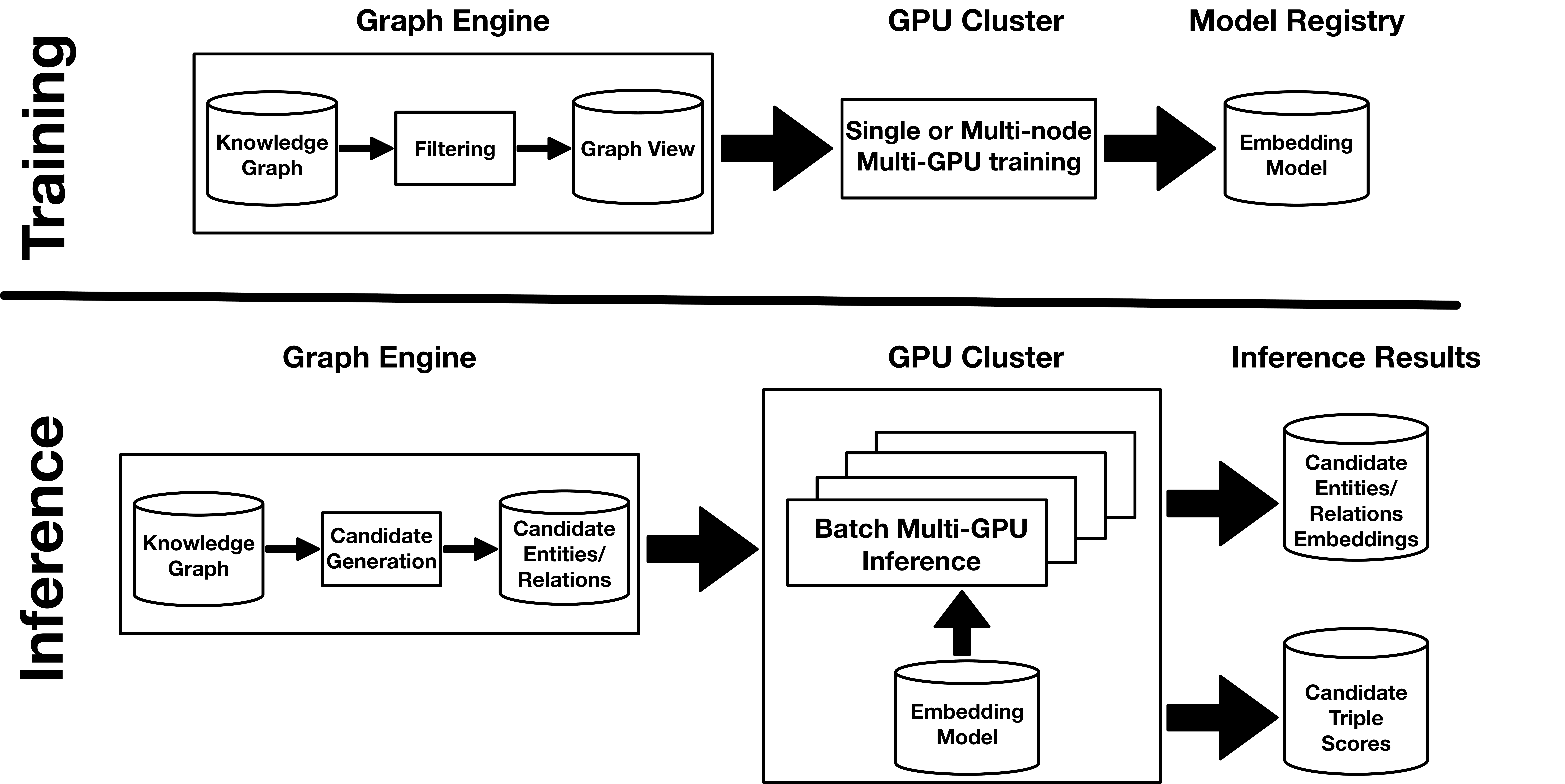}
    \caption{Overview of Embedding Training and Inference}
    \label{fig:embed_overview}
\end{figure}

 In our settings, we consider variations of the last two approaches depending on the models and the downstream task. For general KG embeddings we use disk-based training while for specialized related entity embeddings we use the scalable graph processing capabilities of our graph engine to pre-compute graph traversals. Figure \ref{fig:embed_overview} depicts the overview of the designed system for embedding training and inference. During training, we leverage a computational graph engine to generate a view of the KG by filtering out non-relevant facts and possible noises. Depending on the application, the graph engine might do additional filtering to generate a desired view of triples in the graph. Once we have a filtered view of the graph, depending on the ML model we use for learning the embeddings, we use a single-node or multi-node multi-GPU cluster to train the embedding model. During the inference, similar to the training phase we use the graph engine to generate a set of candidate entities and/or predicates. This set represents candidates corresponding to a specific query. As an example, it might include facts that we need to verify their correctness or even rank them; or it might contain entity pairs for which we need to infer relatedness between them. Once we materialize the candidates, we use a batch inference setting to retrieve embeddings from the learned model and obtain scores for each candidate which denotes its plausibility.

\section{Semantic Annotations} 
\label{sec:sa}
In this section, we present an overview of the ML-based services that use the KG to enable various downstream use-cases e.g., question answering and improved Web search. We collectively call these ``semantic annotation services'' refering to tasks such as, \emph{mention detection}, \emph{named and nominal entity recognition and entity linking}, \emph{contextual entity ranking}, and \emph{searching related entities}. Let us consider an example of Named Entity Recognition and Entity Linking. Our semantic annotation services can disambiguate between two different entities based on the context in which they were mentioned e.g., ``Michael Jordan stats'' should link to KG entity for Michael Jordan the Basketball player while ``Michael Jordan students'' should link to the KG entity for Michael Jordan the professor (please see Figure~\ref{fig:embed_tasks}). This is a very challenging task since lexical similarity-based features alone cannot disambiguate in these scenarios. Such contextual reranking (or disambiguation) between different entities that have names similar to an entity mention in the text can be achieved by computing embeddings on the textual features of the KG entities (e.g., name, description, popularity) and computing a similarity with the query embedding. Such a service can be used to produce more relevant answers in question answering systems or to improve Web document ranking.

\subsection{Linking the Web to KG}
\begin{figure}
    \centering
    \includegraphics[width=0.5\textwidth]{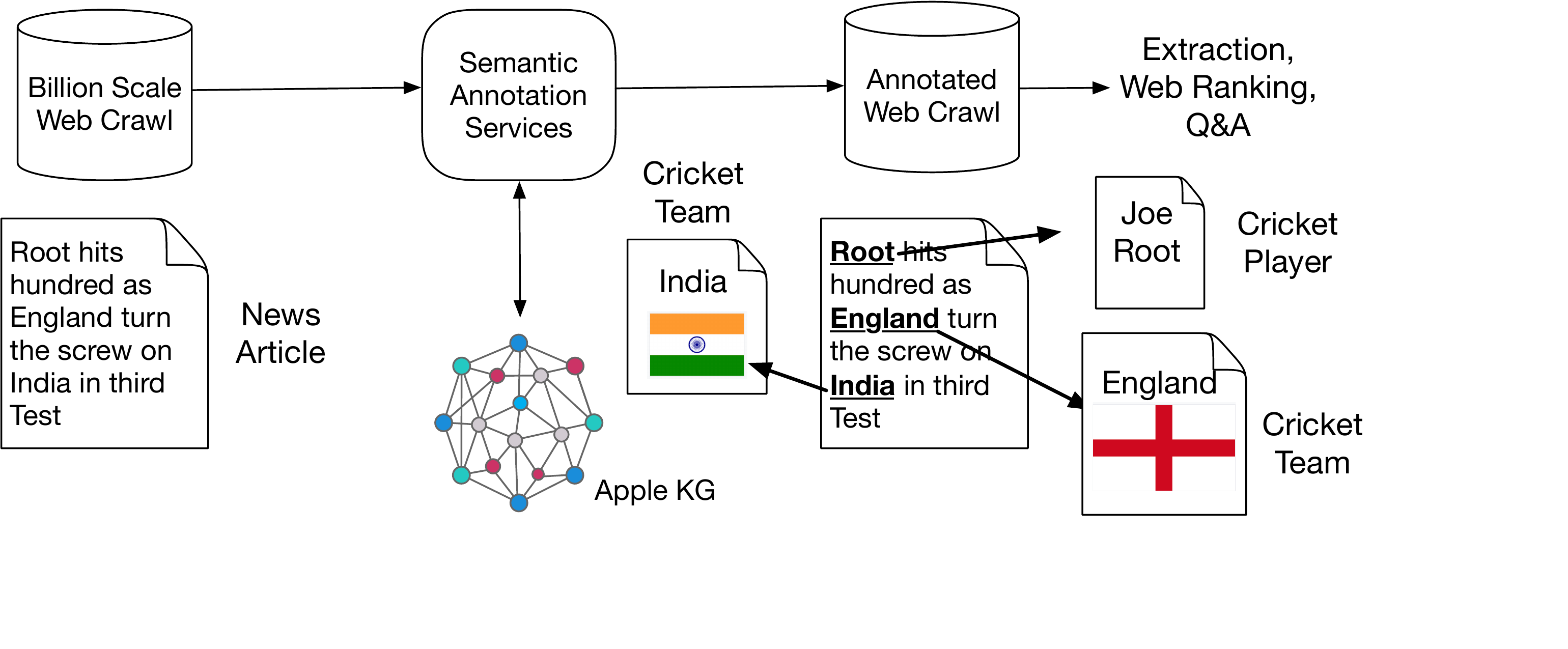}
    \caption{Web-scale Semantic Annotations}
    \label{fig:sa_web}
\end{figure}

The World Wide Web is a large source of structured, semi-structured, and unstructured (textual) data and it has a role in many important user experiences. However, the Web in its raw form has limited utility. To enhance the representation of Web documents, we extend them with semantic annotations from the KG (Figure~\ref{fig:sa_web}). For example, we link all mentions of named or nominal entities to the corresponding KG entities including the corresponding entity types and additional meta-data, such as popularity scores for these entities from various data sources. This enriched view of The Web can be used to improve question answering, ranking related tasks, and for targeted extraction of new facts and entities (see Section~\ref{odke}), among others. However, applying semantic annotations to the whole Web is very challenging for the following reasons:
\begin{itemize}
    \item \textbf{Scale:} The Web contains hundreds of billions of webpages. Our service needs to be able to operate at that scale.
    \item \textbf{Variety:} The Web contains different types of data including, textual, tabular, images and videos. Webpages can also be written in different or mixed languages. Our service needs to be able to handle different data modalities and needs to be multilingual.
    \item \textbf{Rate of change:} The Web is not static. New webpages are constantly created and existing webpages get updated frequently. The service needs to handle incremental changes timely and efficiently.
    \item \textbf{Price/Performance:} Annotating the whole Web is a very costly operation in monetary terms. Our service must be able to navigate the price/performance curve in an effective way i.e., it should provide the best performance with the lowest cost possible.
\end{itemize}

\subsection{An Extensible Semantic Annotations Service}
Given the above challenges, we are implementing a scalable semantic annotation service that can annotate billions of webpages with good price/performance. Our extensible semantic annotation service supports various classical and neural-based components for tasks such as mention detection, named entity recognition, and entity linking with contextual reranking. These services are (1) modular, allowing custom deployments for different use-cases; for example, to balance the requirements for quality (precision and recall) and performance (latency and throughput) to suit the needs of various downstream applications. And (2) are dynamic i.e, are able to surface new and updated entities from the KG, thus ensuring freshness of annotations.  

To improve scalability, we precompute entity embeddings using the methods described in Section~\ref{sec:embeddings} for the contextual reranker and cache the results in a low-latency key-value store. During query time, we only compute the query embeddings and their similarity with the cached entity embeddings. We are also working on extending the service for multiple languages. Our annotation pipeline is able to efficiently process only the changed webpages at a given frequency (e.g., daily or weekly). Last, we are also pushing the boundaries on price/performance by training better and more efficient models using state of the art model distillation and compression techniques that can target different hardware (e.g., CPUs, GPUs, or TPUs) to meet different price/performance SLAs. 

\section{Open Domain Knowledge Extraction} 
\label{odke}

\begin{figure}
    \centering
    \includegraphics[width=0.5\textwidth]{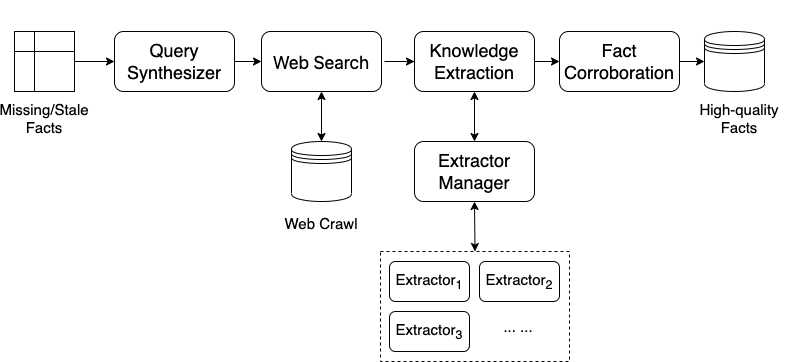}
    \caption{Overview of Open Domain Knowledge Extraction}
    \label{fig:odke_overview}
\end{figure}

An ongoing challenge for any knowledge graph is performing continuous updates which ensure the completeness and freshness of entities and facts in the graph. The coverage and freshness of a knowledge graph directly impacts the quality of downstream applications (e.g. the number of questions that can be correctly answered using the graph). The goal of open domain knowledge extraction (ODKE) is to harvest high-quality facts from the Web to improve the coverage and freshness of a an open domain knowledge graph at scale. ODKE needs to address the following key challenges: 

\begin{itemize}
    \item {\textbf{Volume of data}}. The amount of data and facts contained on the Web is enormous and continuously growing. We need to handle the scalability challenge posed by Web-scale data.   
    \item {\textbf{Variety of data and tasks}}. The Web contains wide varieties of data from plain text to semi-structured data and mixtures of both. Meanwhile, an open domain knowledge graph contains many different types of entities and facts about these entities. In order to extract high-quality facts from the Web, we need to create extractors that are capable of extracting a wide range of high-quality facts for different types of entities from different data sources. 
    \item {\textbf{Veracity}}. The Web is noisy and often contains wrong and conflicting facts. In addition, certain facts, such as someone's marital status or net worth, may also change over time. As such, we need to identify the most accurate and recent facts. 
\end{itemize} 

Figure~\ref{fig:odke_overview} depicts our system designed to address the aforementioned challenges. To address the data volume challenge, we focus on missing and stale facts that matter most and leverage Web search to find relevant documents (Query Synthesizer and Web Search). The important missing/stale facts can be identified in three different ways. First of all, we can reactively identify missing and stale facts by analyzing query logs and finding user queries that are not answered correctly due to missing or stale facts. We can also proactively identify potential coverage and freshness issues within the existing knowledge graph via knowledge graph profiling. In addition, we can predict new facts missing from the current knowledge graph by analyzing potential trending queries or requirements of new use cases. 

\begin{figure}
    \centering
    \includegraphics[width=0.5\textwidth]{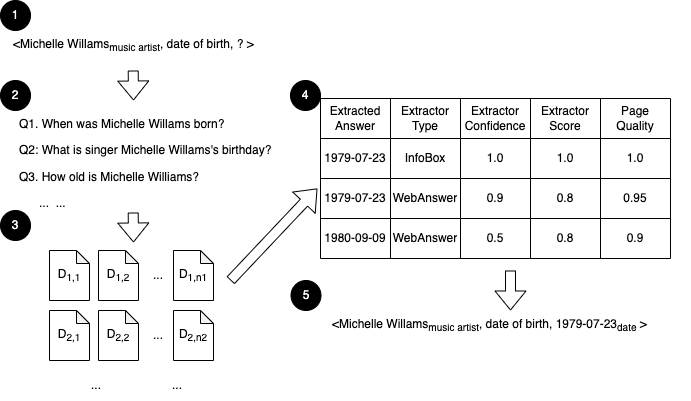}
    \caption{Example for Open Domain Knowledge Extraction: \ding{172} tuple representing a missing fact; \ding{173} auto-generated search queries based on the missing fact; \ding{174} a list of relevant Web documents obtained using the search queries; \ding{175} candidate facts extracted from the documents; \ding{176} final high-quality fact determined based on different evidences and signals.}
    \label{fig:odke_example}
\end{figure}

Important missing or stale facts are given as an input to OKDE. We then use the Query Synthesizer to automatically generate search queries to identify relevant documents that are likely to contain these facts. As illustrated in Figure~\ref{fig:odke_example}, for each missing/stale fact, using a similar approach as~\cite{kamath-etal-2022-improving}, we automatically compose multiple search queries to obtain the relevant documents to gather the fact of interest. Such a targeted search strategy allows us to effectively leverage existing Web search capabilities to significantly reduce the data volume that we need to handle while minimizing impact on the final outcome. 

To overcome the second challenge and properly handle the wide variety of data and tasks that we need to support, we focus on designing different extractors to handle different types of data sources with different types of models. For instance, simple rule-based models can be used to extract key-value pairs from webpages embedded with structured data that conform to schema.org types, while more complex neural models based on large language models are used to extract facts from plain text and leveraging annotations produced by web-scale semantic annotation service as weak labels. In addition, as described in Section~\ref{sec:sa}, the webpages are already linked to the KG via semantic annotation and the annotations can be leveraged to improve retrieval and extraction quality. 

Finally, to address the veracity challenge, we leverage diverse evidence and signals via a trained machine learning model as features to corroborate and identify high quality facts from the list of candidates extracted earlier. For instance, as illustrated by Figure~\ref{fig:odke_example}, after extracting fact candidates from various relevant data sources using the corresponding extractors, we can correctly determine that the missing date of birth for the \emph{music artist} Michelle Williams is {\texttt{1979-07-23}} instead of {\texttt{1980-09-09}} (which is the date of birth for the \emph{actress} Michelle Williams) based on a combination of evidences such as the number of support, extractor type and confidence, and quality of the source page.

\section{On-device Knowledge} 



Personal devices provide a rich source of private and personal information that can be leveraged to personalize on-device experiences such as search, navigation, and recommendations. Because this data is private, all computation must be performed on the users' devices which introduces a number of unique challenges compared to knowledge construction and serving tasks addressed by our server side knowledge platform \cite{saga}. 

\begin{itemize}
    \item Privacy: On-device data sources are private and all computation must happen on the devices themselves.
    \item Resource constrained environments: devices have a wide range of capabilities, and knowledge-based services must be functional within the resource constraints of each hardware environment.
    \item Sync: a user's data ecosystem may span multiple linked devices, knowledge must be kept consistent across devices when allowed or be isolated based on a user's per-source sync preferences.
    \item Global knowledge enrichment: a user's personal knowledge graph needs to be personalized with global knowledge in a privacy preserving way. Global knowledge helps enrich the context of user preferences, such as preferred music genres, news topics, or sports preferences.
\end{itemize}

Our goal for on-device knowledge is to support the full capabilities of the Saga platform, in a private, decentralized, resource constrained environment.

\subsubsection*{Personal KG Construction}
\begin{figure}
    \centering
    \includegraphics[width=0.5\textwidth]{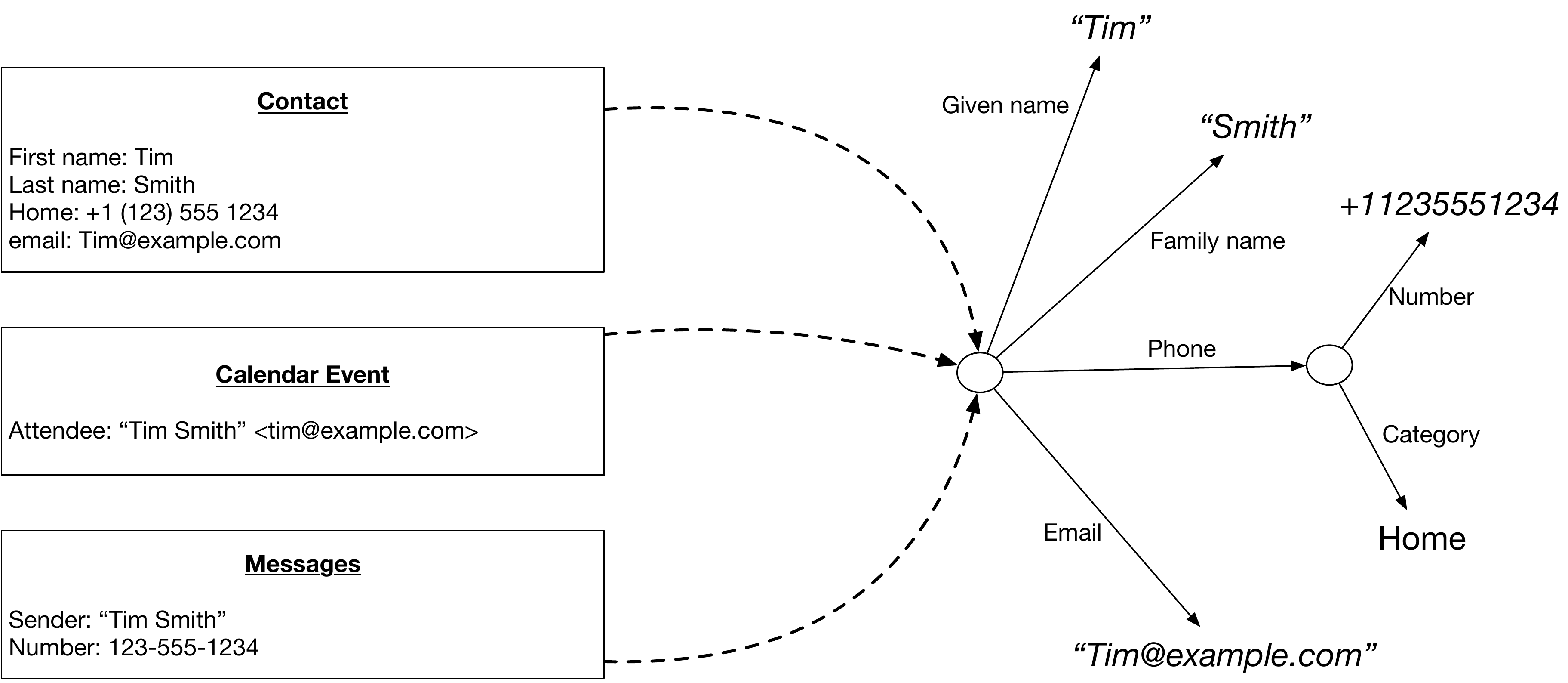}
    \caption{Entity linking for a personal knowledge graph.}
    \label{fig:tim}
\end{figure}

Personal devices have multiple sources of overlapping information that must be linked. For example contact lists, message senders, and calendar invitees all provide sources of \texttt{Person} entities in different formats and namespaces. These sources must be integrated and linked to provide a consolidated representation in a unified ontology. As a motivating example, consider the user utterance \textit{``send a message to Tim''}. There may be a ``Tim'' in the user's contact list, a ``Tim'' that sent the user a text message, and a ``Tim'' that is an invitee on an upcoming calendar event. If we know that the message sender and the contact have the same phone number; that the contact and calendar invitee have the same email address; and that all have similar names; then we may \emph{link} these three source entities into a unified representation of a single \texttt{Person} entity. This allows an utterance understanding algorithm to interpret the reference to ``Tim'' as a reference to a single entity with multiple attributes derived from different sources, rather than having the user disambiguate among three different hypothetical ``Tims.'' Figure~\ref{fig:tim} illustrates this integration scenario. 

This knowledge graph construction task shares many similarities to the graph construction task addressed by our server-side open domain knowledge platform \cite{saga}. However, in an on-device context we must adapt the knowledge platform architecture to accommodate privacy, sync, and resource constraints. 

\subsubsection*{Privacy} 
To ensure all user data is kept private, we build a user's personal knowledge graph entirely on-device. This entails running source ingestion, entity linking, and graph fusion as described in Saga \cite{saga} on the users device. Because resources are limited and highly contended with other on-device background processes, we implement an \emph{incremental continuous construction pipeline}. This pipeline can be paused and resumed at any point without losing state, allowing deferral of the construction process in favor of any other higher priority task (e.g., the user sends a request to the device or another background task is due for scheduling).

\subsubsection*{Sync} 
When possible, a user's personal knowledge graph should be consistent across their devices. This involves privately syncing the KG data in a manner that allows a consistent KG to be constructed on each device. While distributed consistency is a well studied problem, personal KG sync has some unique challenges. In particular, a user may decide to sync or not to sync on a per source basis. For example, a user may sync their contacts and email to all devices, but choose not to sync their calendar. This means the device which contains the calendar source will have additional entities and interactions not present on other devices. However, the sync'd sources still need to be consistently represented across devices. Second, devices have a wide range of compute capabilities (e.g., compare a laptop to a watch). Ensuring a consistent knowledge experience across devices may require offloading expensive computation to more powerful devices, such as expensive views or inference on larger models (see Section 3 of Saga~\cite{saga}), and syncing the result.

\subsubsection*{Resource Constraints} 
To accommodate strict resource constraints that come with on-device processing, we optimize our construction pipeline to be disk oriented with tunable memory buffer sizes. At any given point in the construction pipeline, the amount of memory used is bounded and expensive computations (e.g., pairwise blocking and entity matching described in Section 1 of Saga~\cite{saga}) spill to disk as necessary. Furthermore, we optimize ML models for on-device deployment. On-device ML models are kept small by engineering smaller model architectures (e.g., fewer and more narrow neural layers); compressing learned models (e.g., by floating point precision reduction); or by distillation \cite{distill}. We also leverage CoreML 
to optimize models for our hardware platforms.

\subsubsection*{Semantic Annotation}
Similar to our server-side counterpart, we have a need on device to annotate text with semantic annotations.
Semantic annotation can be used to aid in query and document understanding. As described in Section~\ref{sec:sa}, we also have a need for contextual relevance ranking. As an example, consider a user that has two contacts named ``Tim''. If the user issues the following utterance: \textit{``message Tim that I've added comments to the SIGMOD draft''}, we want to annotate ``Tim'' with the contact most relevant to the query context. In this case, a coworker that has meetings and conversations with the user about ``SIGMOD'' should be ranked above other less relevant contacts named Tim. We achieve this on-device by porting the same Semantic Annotation architecture described in Section 5 of Saga, with smaller models optimized for on-device deployment. This includes a compact neural mention generation and reranker model.

\subsubsection*{Global Knowledge Enrichment}
While a user's personal knowledge graph privately captures context about personal entities and their relationships, it is valuable to complement that with \emph{personalized knowledge}. Information about the user's music, sports, and habitual preferences provide important context that allow the KG to be leveraged for on-device private personalization of experiences. As an example, knowing the typical genre and release year of music the user likes to listen to can help personalize music recommendations. However, querying a server for this information would break user privacy. To address this we extend the on-device knowledge platform with three global knowledge enrichment paths.
\begin{enumerate}
    \item Static knowledge asset: a set of popular entities and facts that are relevant to many users can be sent to every device. As there is no client-side request needed to fetch this artifact, it does not leak any private information. This global knowledge subgraph is implemented as a Graph Engine \emph{view} in Saga. As the set of popular entities changes over time, the view is automatically maintained and can be shipped to devices.
    \item Dynamic enrichment from a user's server interactions: user interactions with global knowledge already involve a request to a server, such as the query \textit{``what is the score in the Blue Jays game?''}. In these scenarios we can piggy-back general knowledge about referenced entities in order to enrich the user's personal graph. Ie., we can include the fact that the \texttt{Blue Jays} are a baseball team located in Toronto.
    \item Private knowledge retrieval: in scenarios where we need global knowledge that is not covered by the static knowledge asset or dynamic enrichment from a user request, we can leverage Differentially Private \cite{dp} knowledge queries or Private Information Retrieval \cite{pir} to fetch global knowledge with provable privacy guarantees. While such approaches are expensive, they can be leveraged for high-value use cases.
\end{enumerate}

\section{Related Work} 

There are two directions related to KG embedding that have been extensively studied. The first is the development of new embedding models such as translational distance embedding models and their generalizations~\cite{zhang2019quaternion, bordes2013translating}, semantic matching embedding models~\cite{yang2014embedding}, and reasoning-based models~\cite{ren2020query2box}. The second direction is learning embeddings at scale. Due to the significant growth in the size of KGs in recent years, there is an active line of research on designing systems that can handle large-scale training of KG embeddings, such as Marius~\cite{mohoney2021marius, mariusgnn}, Pytorch Biggraph~\cite{lerer2019pytorch}, and DGL-KE~\cite{zheng2020dgl}.

Various semantic annotation tasks, including entity linking and disambiguation, have received renewed interest from the research community, especially for deep learning based approaches ~\cite{e2e-neural-el-2018, deep-joint-el-2017, latent-relations-el-2018, deep-type-2018}. Prior works have also discussed the applications of semantic annotations to enrich webpages for improved question answering and web search ~\cite{balog2018entity}. 

Harvesting knowledge from Internet sources is a well-studied problem ~\cite{DBLP:series/lncs/WeikumHS19}. Many existing knowledge bases, such as Yago (\url{yago-knowledge.org}), Google Knowledge Graph~\cite{DBLP:conf/kdd/0001GHHLMSSZ14}, Amazon Product Knowledge Graph~\cite{DBLP:conf/kdd/Dong18}, IBM Financial Content Knowledge Base~\cite{DBLP:journals/pvldb/BharadwajCDDDCG17} and many others, are at least partially automatically constructed based on Internet sources. Many efforts are based on distantly supervised approach, centered on the principle of pattern-fact duality. The most popular source for knowledge harvesting is Wikipedia, including semi-structured elements such as InfoBox as well as plain text. Web-scale resources have also been exploited but to a much lesser degree and mostly focusing on extraction from tables~\cite{DBLP:conf/kdd/0001GHHLMSSZ14}. We are not aware of any published work that focus on web-scale extraction on top of web-scale semantic annotation and leverage wide variety of sources with more sophisticated extractors such as ones based on neural networks.

\section{Conclusion}
We have presented important extensions to our knowledge platform, Saga. 
Graph embeddings form a core capability on which we build enabling features such as fact ranking, fact verification, a related entities service, and support for entity linking. In particular, entity linking can be leveraged as part of a general semantic annotation service. We have described how this service can be leveraged to link Web documents to our KG, extending our notion of a knowledge graph to include links from entities to unstructured documents. Finally, we have described how our KG grow with targeted fact extraction using Open Domain Knowledge Extraction. 

We also presented adaptations to our knowledge platform needed to construct and serve private personal knowledge on-device. This includes the unique challenges of private incremental KG construction, knowledge graph sync, and global knowledge enrichment.


\subsection*{ACKNOWLEDGEMENTS}
This work was made possible by the hard work of many people at Apple. We would like to thank the entire Knowledge Platform and Intelligence Platform teams for their many contributions.

\bibliographystyle{ACM-Reference-Format}
\balance
\bibliography{kp_paper.bib}

\end{document}